\documentclass[a4paper,fleqn]{cas-dc}

\usepackage[authoryear]{natbib}
\usepackage{amsmath}
\usepackage{amssymb}
\usepackage{booktabs}
\usepackage{multirow}
\usepackage{bm}
\usepackage{algorithm}
\usepackage{algorithmicx}
\usepackage{algpseudocode}
\usepackage{subcaption}
\usepackage[table]{xcolor}
\usepackage{graphicx}

\def\tsc#1{\csdef{#1}{\textsc{\lowercase{#1}}\xspace}}
\tsc{WGM}
\tsc{QE}

\begin{document}
\let\WriteBookmarks\relax
\def\floatpagepagefraction{1}
\def\textpagefraction{.001}

\shorttitle{}    

\shortauthors{}  

\title[mode = title]{DLaR: Dynamic Stochasticity Control for Latent Reasoning in Large Language Models}  

% % Title footnote mark

% % Title footnote 1.

\author[suda]{Xiaocheng Luo}
\ead{xcluo@stu.suda.edu.cn}

\author[suda]{Peng Xu}
\ead{pxu26@stu.suda.edu.cn}

\author[ece]{Zaifu Zhan}
\ead{zhan8023@umn.edu}

\author[suda]{Xiangyu Duan}
\ead{xiangyuduan@suda.edu.cn}
\cormark[1]
\cortext[1]{Corresponding author}

\affiliation[suda]{organization={School of Computer Science and Technology},
    addressline={Soochow University}, 
    city={Suzhou},
    postcode={215006}, 
    state={Jiangsu},
    country={PR China}}

\affiliation[ece]{organization={Department of Electrical and Computer Engineering},
    addressline={University of Minnesota}, 
    city={Minneapolis},
    postcode={55455}, 
    state={MN},
    country={USA}}

% Abstract
\begin{abstract}
    Latent reasoning enables LLMs to operate in continuous embedding space, yet existing methods suffer from deterministic inference that precludes exploration of diverse reasoning paths. While recent approaches inject stochasticity to restore exploration, we identify two fundamental limitations: \textit{uniform stochasticity} across all steps is suboptimal, and unconstrained perturbations disrupt reasoning coherence. We propose \textbf{D}ynamic Stochasticity Control for \textbf{La}tent \textbf{R}easoning (\textbf{DLaR}), a framework with two complementary mechanisms that operate at different temporal granularities within the GRPO reinforcement learning framework. At the token level, \textbf{Adaptive Stochasticity Modulation (ASM)} dynamically adjusts perturbation intensity at each latent step based on a confidence-entropy fusion signal, increasing exploration when the model is uncertain and preserving stability when confident. At the training level, \textbf{Training-Aware Stochasticity Tuning (TAST)} progressively decays the base stochasticity level following a curriculum schedule. Through this dual-level design, DLaR establishes a hierarchical exploration-exploitation strategy that operates at both token and training levels. Extensive experiments on four in-distribution and three out-of-distribution mathematical reasoning benchmarks demonstrate that DLaR achieves the highest average performance across model scales, reaching 27.9\%/50.6\% Pass@1/Pass@8 on Llama-3.2-1B and 46.8\%/68.2\% on Qwen2.5-1.5B. Code: \href{https://github.com/xiaocen-luo/DLaR}{https://github.com/xiaocen-luo/DLaR}.
\end{abstract}

\begin{highlights}
\item Uniform stochasticity across reasoning steps is suboptimal.
\item Unconstrained perturbations disrupt latent reasoning coherence.
\item DLaR adapts perturbation via confidence-entropy and curriculum decay.
\item DLaR achieves the highest average on math reasoning benchmarks.
\end{highlights}

% Keywords

\maketitle

% Main text begins
\section{Introduction}
\label{sec:intro}

Large Language Models (LLMs) have demonstrated remarkable reasoning capabilities through Chain-of-Thought (CoT) prompting \citep{wei2022chain} and reinforcement learning-based optimization \citep{shao2024deepseekmath, guo2025deepseek}. However, discrete token-level reasoning suffers from inherent information loss, as each step must commit to a single token, discarding the rich distributional information encoded in the model's internal states. To address this limitation, recent work has shifted toward \textbf{latent reasoning} \citep{wang2025latent, zhang2025soft, kang2026improving, luo2026ruplarefficientlatent}, where reasoning occurs in continuous embedding space, preserving richer semantic information and enabling more comprehensive reasoning \citep{hao2025training}.

Despite these advantages, existing latent reasoning methods face a critical limitation: \textbf{deterministic inference}. Without stochastic sampling, the latent reasoning process produces an invariant output for a given input, collapsing to deterministic trajectories that fail to explore diverse reasoning paths. This lack of exploration significantly hinders the model's ability to discover correct solutions, particularly for complex mathematical problems that require multi-step reasoning.

Recent approaches have introduced stochasticity into latent reasoning to restore exploratory capacity, employing techniques such as Gumbel-Softmax \citep{zhou2026lepo, zheng2025soft}, Gaussian noise \citep{butt2026soft}, or Dirichlet perturbations \citep{wu2026llms}. While conceptually appealing, we identify two fundamental limitations with existing approaches that inject stochasticity into latent reasoning, which critically undermine their effectiveness.

\begin{figure*}
    \centering
    \begin{subfigure}[b]{0.34\linewidth}
        \centering
        \includegraphics[width=\textwidth]{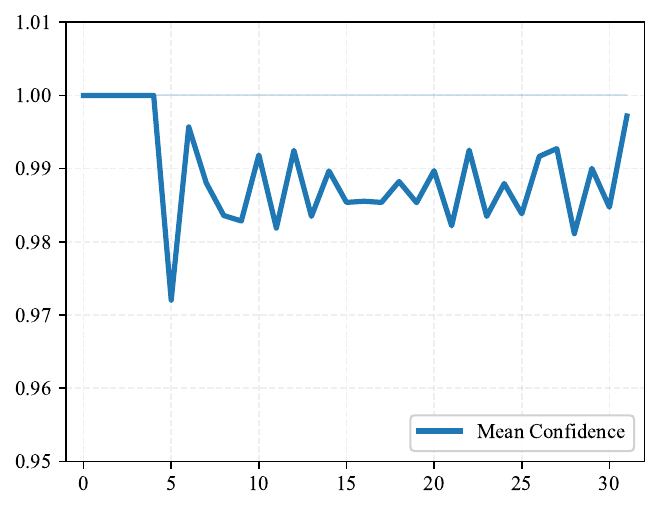}
        \caption{}
        \label{fig:confidence_dist}
    \end{subfigure}
    \hfill
    \begin{subfigure}[b]{0.34\linewidth}
        \centering
        \includegraphics[width=\textwidth]{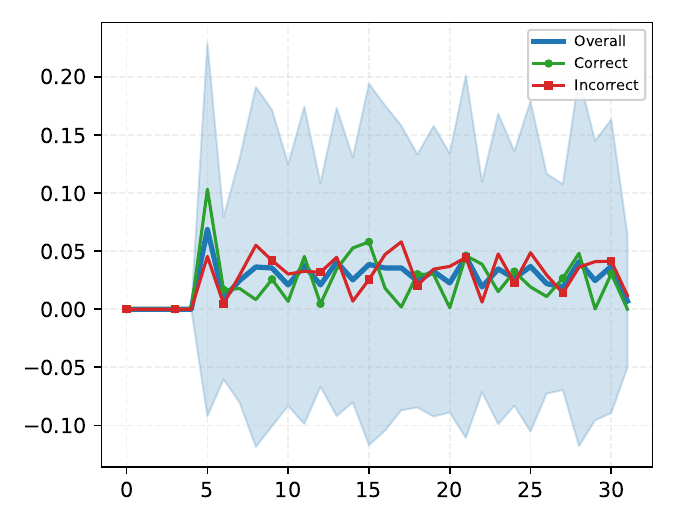}
        \caption{}
        \label{fig:entropy_curves}
    \end{subfigure}
    \hfill
    \begin{subfigure}[b]{0.29\linewidth}
        \centering
        \includegraphics[width=\textwidth]{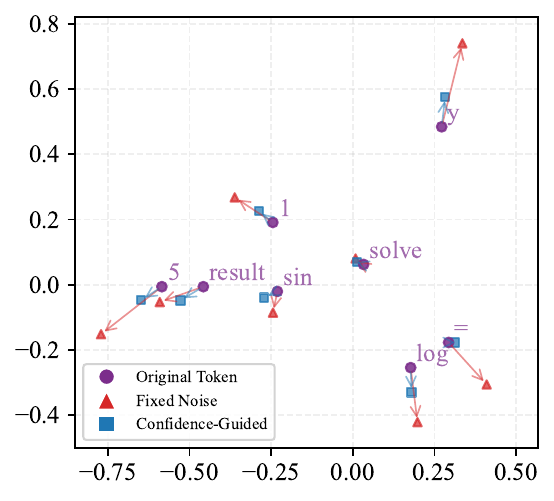}
        \caption{}
        \label{fig:noise_compare}
    \end{subfigure}
    \caption{Empirical motivation for dynamic stochasticity control. (a) Confidence remains uniformly high across latent steps. (b) Correct paths maintain higher entropy than incorrect ones. (c) Fixed perturbation causes semantic drift, while adaptive perturbation preserves coherence.}
    \label{fig:intro_overview}
\end{figure*}

\textbf{First, uniform stochasticity across all reasoning steps is suboptimal.} To examine this, we sample 8 trajectories per question on GSM8K using a DLaR-trained Llama-3.2-1B model and label each path as correct or incorrect based on exact answer match. As shown in Figure~\ref{fig:confidence_dist}, confidence remains uniformly high (>0.97) across all latent steps—the model is overconfident and lacks inherent exploration. Figure~\ref{fig:entropy_curves} further shows that correct paths maintain higher entropy than incorrect ones, especially during middle stages, suggesting that sustained uncertainty is critical for successful reasoning. These observations indicate that confidence and entropy provide complementary signals for guiding adaptive stochasticity control: confidence reveals where the model is overconfident, while entropy indicates whether the model is on the right track.

\textbf{Second, unconstrained perturbations can disrupt reasoning coherence across model scales.} While the effect is particularly pronounced in smaller models due to their capacity constraints, we argue that this is a general problem affecting models of all sizes. The disruption manifests as \textbf{semantic drift}, where injected noise pushes latent representations \textit{off the semantic manifold}, leading to incoherent reasoning paths. Figure~\ref{fig:noise_compare} contrasts unconstrained perturbation injection, which causes significant \textit{off-manifold} shifts (red triangles), with adaptive perturbation (blue square), which preserves semantic coherence by adjusting intensity based on a confidence-entropy fusion signal.

These two limitations, if left unaddressed, lead to a fundamental dilemma for stochasticity in latent reasoning: uniform perturbation either over-explores on confident steps—potentially disrupting stable reasoning—or under-explores when exploration is most needed, while unconstrained noise introduces semantic drift that undermines reasoning coherence. As a result, existing methods cannot fully exploit the potential of latent reasoning, particularly on complex mathematical problems where reliable reasoning paths are critical.

To address these limitations, we propose \textbf{DLaR}, a framework that introduces dynamic stochasticity control for latent reasoning. Our framework is built upon two key insights. First, different reasoning steps require different levels of exploration: confident steps benefit from reduced perturbation, while uncertain steps benefit from increased stochasticity. Second, the overall perturbation level should follow a curriculum decay throughout training, enabling broad exploration in early stages and fine-grained exploitation later as reliable reasoning patterns emerge.

Building on these insights, DLaR comprises two complementary mechanisms that operate at different temporal granularities. At the token level, \textbf{ASM} dynamically adjusts perturbation intensity at each latent step based on a confidence-entropy fusion signal, increasing exploration when the model is uncertain and preserving stability when confident. At the training level, \textbf{TAST} progressively decays the base stochasticity level following a curriculum schedule. These mechanisms establish a hierarchical exploration-exploitation strategy that controls stochasticity at both token and training levels. Experiments on four in-distribution mathematical reasoning benchmarks and three out-of-distribution generalization benchmarks demonstrate that DLaR achieves the highest average performance across model scales.

Our contributions are summarized as follows:
\begin{itemize}
    \item We identify and systematically analyze two fundamental limitations of existing approaches that inject stochasticity into latent reasoning: uniform stochasticity across all steps is suboptimal, and unconstrained perturbations disrupt reasoning coherence.
    \item We propose \textbf{DLaR}, a dynamic stochasticity control framework with two complementary mechanisms: \textbf{ASM}, which modulates perturbation intensity at each token step via confidence-entropy fusion, and \textbf{TAST}, which progressively decays the base stochasticity level throughout training.
    \item Extensive experiments on in-distribution and out-of-distribution mathematical reasoning benchmarks demonstrate that DLaR achieves the highest average performance across model scales. Our analysis further reveals that DLaR maintains appropriate entropy levels throughout training, enabling sustained exploration without compromising reasoning quality.
\end{itemize}

\section{Related Work}
\label{sec:related}

\subsection{Latent Reasoning}

Traditional reasoning in Large Language Models relies on Chain-of-Thought \citep{wei2022chain}, where intermediate reasoning steps are expressed as discrete tokens. However, this discrete nature introduces information loss, as each step must collapse the model's distributional beliefs into a single token.

Recent work has explored \textbf{latent reasoning}, where reasoning occurs in continuous embedding space rather than discrete token space. \citet{hao2025training} propose Coconut, which replaces discrete CoT tokens with continuous hidden states, enabling reasoning in a more information-rich space. \citet{zhang2025soft} introduce Soft Thinking, which uses weighted combinations of token embeddings as latent representations. \citet{zhu2025reasoning} provide a theoretical analysis of latent reasoning, demonstrating its advantages over discrete reasoning in terms of information preservation. A key limitation of these early methods is \textbf{deterministic inference}—the reasoning process produces invariant outputs, lacking the stochastic exploration necessary for discovering diverse reasoning paths.

\subsection{Stochasticity in LLM Reasoning}

To address this limitation, several recent approaches have introduced stochasticity into latent reasoning. \citet{wu2026llms} first demonstrated that stochastic perturbations—via Dirichlet noise or Gumbel-Softmax—can enhance LLM reasoning by preventing deterministic collapse. Building on this insight, subsequent work has explored various perturbation injection strategies.

\citet{zhou2026lepo} propose LEPO, which injects controllable stochasticity via Gumbel-Softmax and optimizes the latent policy with reinforcement learning. \citet{zheng2025soft} propose Soft-GRPO, which applies Gumbel reparameterization to logits for soft-thinking policy optimization. \citet{butt2026soft} inject Gaussian noise directly into token embeddings or logits during rollout. Despite their effectiveness, these perturbation-based methods apply uniform perturbation intensity across all steps and do not adapt to the model's evolving confidence—precisely the limitations we address in this work.

Beyond perturbation injection, \citet{zhu2026ngrpo} propose Semantic Neighbor Mixing, which constrains perturbations to the semantic manifold to prevent semantic drift—a complementary direction that also relates to reasoning coherence. However, it requires pre-computed neighbor sets and does not adapt to model confidence.

\section{Method}
\label{sec:method}

In this section, we present \textbf{DLaR}, a novel framework that introduces dynamic stochasticity control for latent reasoning. Our method is built upon two key insights: (1) different reasoning stages require different levels of stochastic exploration, which can be guided by a confidence-entropy fusion signal; and (2) the overall perturbation level should decrease progressively during training following a curriculum schedule.

\begin{figure*}
    \centering
    \includegraphics[width=1.0\linewidth]{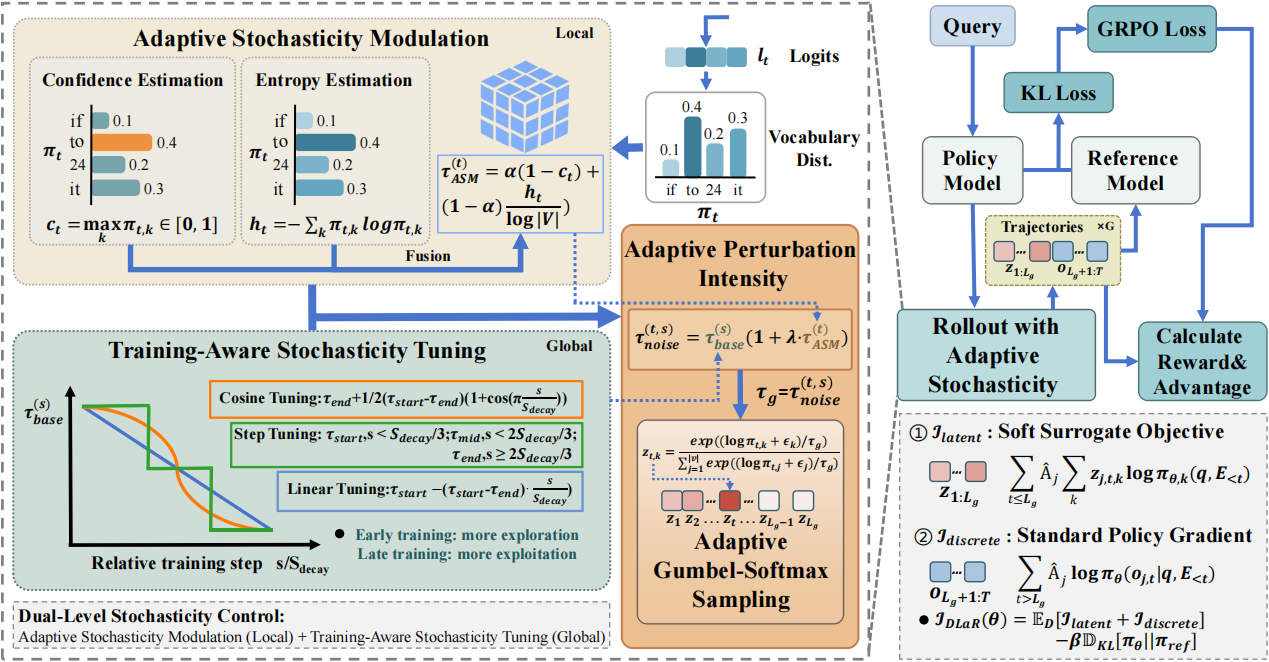}
    \caption{\label{fig:our_model} Overview of the DLaR framework. The rollout stage samples trajectories with adaptive perturbation intensities, while the optimization stage computes unified gradients for both latent and discrete tokens. GRPO serves as the underlying RL framework.}
\end{figure*}

As illustrated in Figure~\ref{fig:our_model}, DLaR consists of two complementary mechanisms operating at different temporal granularities: (1) \textbf{ASM}, a local, token-level mechanism that dynamically adjusts perturbation intensity at each latent step based on a confidence-entropy fusion signal; and (2) \textbf{TAST}, a global, training-level mechanism that progressively tunes the base stochasticity level throughout training in accordance with the model's learning progress. These components enable fine-grained control over the exploration-exploitation trade-off at both the step level and the training level.

\subsection{Preliminaries: Latent Reasoning with Stochasticity}
\label{subsec:preliminaries}

We first review the latent reasoning framework that serves as the foundation for our method. Given an input query $q$, the model generates a sequence of \textbf{latent tokens} $\{\mathbf{z}_1, \ldots, \mathbf{z}_{L_g}\}$ in continuous space, followed by discrete answer tokens $\{o_{L_g+1}, \ldots, o_T\}$, where $o_t$ denotes the discrete token sampled at step $t$.

At each latent step $t$ ($t \leq L_g$), the model produces a vocabulary distribution over the original logits. To enable stochastic exploration, existing approaches apply perturbation injection techniques such as Gumbel-Softmax \citep{zhou2026lepo}, Gaussian noise \citep{butt2026soft}, or Dirichlet perturbations \citep{wu2026llms}. Taking Gumbel-Softmax as an example, the stochastic latent token is obtained as:

\begin{equation}
    \mathbf{z}_{t,k} = \frac{\exp((\log \pi_{t,k} + \epsilon_k) / \tau_g)}{\sum_{j=1}^{|\mathcal{V}|} \exp((\log \pi_{t,j} + \epsilon_j) / \tau_g)},
    \label{eq:gumbel}
\end{equation}

\noindent where $\epsilon_k \sim \text{Gumbel}(0,1)$ is independent Gumbel noise, $\tau_g$ is the temperature controlling the sharpness of the distribution, and $\pi_t$ is the original vocabulary distribution from the model.

The input embedding for the next step is computed as the expectation over token embeddings:

\begin{equation}
    \mathbf{E}_t = \sum_{k=1}^{|\mathcal{V}|} \mathbf{z}_{t,k} \cdot \mathbf{e}_k,
    \label{eq:embedding}
\end{equation}

\noindent where $\mathbf{e}_k$ is the embedding of the $k$-th vocabulary token.

\subsection{Adaptive Stochasticity Modulation}
\label{subsec:asm}

Existing approaches that inject stochasticity into latent reasoning apply a fixed perturbation intensity uniformly across all latent steps. We argue that this uniform strategy is fundamentally suboptimal, as different reasoning stages require different levels of exploration. To address this, we propose \textbf{ASM}, a local mechanism that dynamically adjusts perturbation intensity at each latent step based on a confidence-entropy fusion signal. The core insight is that confidence and entropy provide complementary signals: confidence reflects the model's immediate predictive certainty, while entropy captures broader distributional uncertainty. Together, they enable a principled exploration-exploitation trade-off at the token level. ASM consists of two sequential components: confidence estimation quantifies the model's predictive certainty, entropy captures distributional uncertainty, and their fusion produces a robust modulation signal for stochasticity adaptation.

\subsubsection{Confidence Estimation}

At each latent step $t$, the model produces the logits vector $\mathbf{l}_t \in \mathbb{R}^{|\mathcal{V}|}$, which is scaled by the sampling temperature $\tau$ to obtain the vocabulary distribution:

\begin{equation}
    \pi_t = \text{softmax}\left(\frac{\mathbf{l}_t}{\tau}\right),
    \label{eq:distribution}
\end{equation}

\noindent where $\tau$ is the sampling temperature. We define the model's confidence as the maximum probability:

\begin{equation}
    c_t = \max_{k} \pi_{t,k} \in [0, 1],
    \label{eq:confidence}
\end{equation}

To capture distributional uncertainty more comprehensively, we additionally compute the entropy:

\begin{equation}
    h_t = -\sum_{k} \pi_{t,k} \log \pi_{t,k},
    \label{eq:entropy}
\end{equation}

High confidence (low entropy) indicates the model is certain about its current reasoning direction, while low confidence (high entropy) suggests uncertainty and the need for exploration.

\subsubsection{Confidence-Entropy Fusion}

While confidence alone provides a natural signal for modulating stochasticity, entropy captures complementary distributional information that reflects the model's broader uncertainty. We therefore combine both measures to define the ASM modulation signal:

\begin{equation}
    \tau_{\text{ASM}}^{(t)} = \alpha (1 - c_t) + (1-\alpha) \frac{h_t}{\log |\mathcal{V}|},
    \label{eq:asm_signal}
\end{equation}

\noindent where $\alpha \in [0, 1]$ controls the relative importance of confidence vs. entropy, and $\tau_{\text{ASM}}^{(t)} \in [0, 1]$ quantifies the model's uncertainty at step $t$, with higher values indicating greater need for exploration. When $\alpha = 1$, the modulation relies solely on confidence; when $\alpha = 0$, it relies solely on normalized entropy. We evaluate the impact of different $\alpha$ values in our hyperparameter analysis (Appendix~\ref{app:hyperparameter}).

\subsection{Training-Aware Stochasticity Tuning}
\label{subsec:tast}

While ASM provides fine-grained, step-wise stochasticity adaptation, we additionally introduce \textbf{TAST}, a global mechanism that progressively decays the base stochasticity level throughout training in accordance with the model's learning progress. This follows the intuition that the model requires broad exploration in early training stages when prior knowledge is limited, and fine-grained exploitation in later stages when reliable reasoning patterns have been established. TAST operates at the training level, complementing ASM's token-level adaptation to form a hierarchical stochasticity control strategy.

We implement three tuning schedules:

\begin{enumerate}
    \item \textbf{Linear Tuning}: 
    \begin{equation}
        \tau_{\text{base}}^{(s)} = \tau_{\text{start}} - (\tau_{\text{start}} - \tau_{\text{end}}) \cdot \dfrac{s}{S_{\text{decay}}},
        \label{eq:linear_decay}
    \end{equation}

    \item \textbf{Cosine Tuning}: 
    \begin{equation}
        \small
        \tau_{\text{base}}^{(s)} = \tau_{\text{end}} + \dfrac{1}{2}(\tau_{\text{start}} - \tau_{\text{end}})\left(1 + \cos\left(\pi \dfrac{s}{S_{\text{decay}}}\right)\right),
        \label{eq:cosine_decay}
    \end{equation}

    \item \textbf{Step Tuning}: 
    \begin{equation}
        \tau_{\text{base}}^{(s)} = \begin{cases} 
        \tau_{\text{start}}, & s < S_{\text{decay}}/3 \\
        \tau_{\text{mid}}, & S_{\text{decay}}/3 \leq s < 2S_{\text{decay}}/3 \\
        \tau_{\text{end}}, & s \geq 2S_{\text{decay}}/3 
    \end{cases},
        \label{eq:stepwise_decay}
    \end{equation}
\end{enumerate}

\noindent where $s$ denotes the current step within the decay phase, $S_{\text{decay}}$ is the total number of decay steps, and $\tau_{\text{start}}, \tau_{\text{mid}}, \tau_{\text{end}}$ are the start, middle, and end stochasticity levels. These schedules offer different annealing profiles: linear tuning provides a steady and predictable reduction, cosine tuning offers a smooth decay with an initial acceleration, and step tuning introduces discrete phase transitions that align with distinct training stages. We compare these schedules in our experiments (Appendix~\ref{app:hyperparameter}).

\subsection{Adaptive Perturbation Intensity}
\label{subsec:adaptive_perturbation}

Combining ASM and TAST, the final perturbation intensity at step $t$ of training step $s$ is:

\begin{equation}
    \small
    \tau_{\text{noise}}^{(t,s)} = \underbrace{\tau_{\text{base}}^{(s)}}_{\text{TAST}} \cdot (1 + \lambda \cdot \underbrace{\tau_{\text{ASM}}^{(t)}}_{\text{ASM}}),
    \label{eq:final_noise}
\end{equation}

\noindent where $\tau_{\text{base}}^{(s)}$ is the base stochasticity level scheduled by TAST, $\tau_{\text{ASM}}^{(t)}$ is the modulation signal from Equation~(\ref{eq:asm_signal}), and $\lambda$ controls the influence of ASM on the final perturbation intensity. This dual-level adaptation ensures that stochasticity is controlled globally (across training steps via TAST) and locally (across reasoning steps via ASM), establishing a hierarchical exploration-exploitation strategy that operates at both token and training levels.

The resulting $\tau_{\text{noise}}^{(t,s)}$ replaces the fixed temperature $\tau_g$ in Equation~(\ref{eq:gumbel}), so that the latent token at each step is sampled as:

\begin{equation}
    \mathbf{z}_{t,k} = \frac{\exp((\log \pi_{t,k} + \epsilon_k) / \tau_{\text{noise}}^{(t,s)})}{\sum_{j=1}^{|\mathcal{V}|} \exp((\log \pi_{t,j} + \epsilon_j) / \tau_{\text{noise}}^{(t,s)})},
    \label{eq:adaptive_gumbel}
\end{equation}

\noindent where $\epsilon_k \sim \text{Gumbel}(0,1)$ is independent Gumbel noise. This adaptive sampling formulation allows each latent token to be generated with a perturbation intensity that reflects the model's current uncertainty at both the token and training levels.

\subsection{Integration with Reinforcement Learning}
\label{subsec:integration}

We integrate our adaptive stochasticity control mechanisms into the GRPO \citep{shao2024deepseekmath} framework for latent reasoning optimization. GRPO eliminates the need for a separate value network, reducing memory and computational overhead compared to PPO, and has proven effective for reasoning tasks \citep{guo2025deepseek}. The overall framework consists of a rollout stage and an optimization stage.

\subsubsection{Rollout Stage}

In the rollout stage, we generate $G$ trajectories for each query $q$ by sampling from the current policy. For each trajectory $j$ ($j = 1, \ldots, G$), at each latent step $t$, the vocabulary distribution $\pi_{j,t}$ is obtained from the model. We then compute the confidence $c_{j,t}$ and entropy $h_{j,t}$ from this distribution, which are used to determine the ASM modulation signal $\tau_{\text{ASM}}^{(t)}$ for trajectory $j$ as defined in Equation~(\ref{eq:asm_signal}). The final perturbation intensity $\tau_{\text{noise}}^{(t,s)}$ is then obtained by scaling with the current base stochasticity level $\tau_{\text{base}}^{(s)}$ from TAST, as defined in Equation~(\ref{eq:final_noise}). The latent token is sampled via Gumbel-Softmax using this adaptive perturbation intensity, and the generation proceeds autoregressively. This design allows each trajectory to balance exploration and exploitation autonomously: trajectories with high confidence proceed with minimal perturbation to preserve stable reasoning, whereas those with low confidence leverage increased stochasticity to explore alternative reasoning paths.

\subsubsection{Optimization Stage}

We adopt the unified optimization objective that supervises both latent and discrete tokens. For each trajectory $j$, let $o_{j,t}$ denote the discrete token sampled at step $t$ (for $t > L_g$), and $\mathbf{z}_{j,t,k}$ denote the $k$-th component of the latent token sampled at step $t$ (for $t \leq L_g$). The gradient is computed over both types of tokens:

\begin{equation}
    \begin{aligned}
        \mathcal{J}_{\text{total}}^{(j)} 
        &= \sum_{t \leq L_g} \hat{A}_j \sum_{k} \mathbf{z}_{j,t,k} \log \pi_{\theta,k}(\mathbf{q}, \mathbf{E}_{<t}) \\
        &\quad + \sum_{t > L_g} \hat{A}_j \log \pi_{\theta}(o_{j,t} \mid \mathbf{q}, \mathbf{E}_{<t}),
    \end{aligned}
    \label{eq:objective}
\end{equation}

\noindent where $\hat{A}_j$ is the group-relative advantage computed from the rewards. The first term provides supervision to latent tokens through a soft surrogate objective, while the second term applies the standard policy gradient to discrete tokens.

The overall optimization objective is:

\begin{equation}
    \mathcal{J}_{\text{DLaR}}(\theta) = \mathbb{E}_{\mathcal{D}}\left[\frac{1}{G} \sum_{j=1}^{G} \frac{1}{T_j} \mathcal{J}_{\text{total}}^{(j)}\right] - \beta \mathbb{D}_{\text{KL}}[\pi_{\theta} \| \pi_{\text{ref}}],
    \label{eq:dlr_objective}
\end{equation}

\noindent where $T_j$ is the length of the $j$-th trajectory, $\pi_{\text{ref}}$ is the reference model for KL-divergence regularization, and $\beta$ is the coefficient controlling the strength of the KL penalty.

\section{Experiments}

\subsection{Experimental Setup}

\textbf{Datasets.}
We evaluate on four in-distribution mathematical reasoning benchmarks: GSM8K \citep{cobbe2021training}, MATH500 \citep{hendrycks2021measuring}, MinervaMath \citep{lewkowycz2022minerva}, and AMC23 \citep{he2026deepmath}. For out-of-distribution evaluation, we use ARC-C \citep{clark2018think}, GPQA \citep{rein2024gpqa}, and MMLU-STEM \citep{hendrycks2021measuring}. All models are trained on the DAPO-Math-17k dataset \citep{yu2026dapo}.

\textbf{Baseline Methods.}
We compare DLaR against three categories of baselines. For discrete reasoning, we include \textbf{GRPO} \citep{shao2024deepseekmath}, which operates on discrete tokens without latent reasoning. For training-free latent reasoning, we include \textbf{Soft Thinking} \citep{wu2026llms} with both Gumbel and Dirichlet noise, applying stochasticity at inference time without parameter updates. For RL-trained latent reasoning, we include \textbf{Soft Tokens, Hard Truths} \citep{butt2026soft}, which injects Gaussian noise during rollout, and \textbf{LEPO} \citep{zhou2026lepo}, which applies Gumbel-Softmax for controllable stochasticity. 

\textbf{Implementation Details.}
We use GRPO \citep{shao2024deepseekmath} with group size $G=8$ and batch size 1. For DLaR, we set $L_g=32$, $\tau_{\text{start}}=0.35$, $\tau_{\text{end}}=0.02$, $\lambda=0.5$, and $\alpha=0.7$, using Linear Tuning for the base stochasticity schedule. All methods are trained under identical settings---Llama-3.2-1B for 1 day and Qwen2.5-1.5B for 1.5 days---with the best checkpoint reported. More implementation details are provided in Appendix~\ref{app:implementation_details}.

\subsection{Results}

\begin{table*}[!ht]
    \centering
    \small
    \setlength{\tabcolsep}{2pt}
    \resizebox{\linewidth}{!}{
        \begin{tabular}{l c c c c c c c c| c c}
            \toprule
            \multirow{2}{*}{Model} & \multicolumn{2}{c}{GSM8K} & \multicolumn{2}{c}{MATH500} & \multicolumn{2}{c}{MinervaMath} & \multicolumn{2}{c}{AMC23} & \multicolumn{2}{|c}{Average} \\
            \cmidrule(lr){2-3} \cmidrule(lr){4-5} \cmidrule(lr){6-7} \cmidrule(lr){8-9} \cmidrule(lr){10-11}
            & Pass@1 & Pass@8 & Pass@1 & Pass@8 & Pass@1 & Pass@8 & Pass@1 & Pass@8 & Pass@1 & Pass@8 \\
            \hline
            \multicolumn{11}{l}{\cellcolor{cyan!20}\textbf{Llama-3.2-1B}} \\
            \hline
            \multicolumn{11}{l}{\textbf{Discrete Reasoning}} \\
            \quad Base Model    & 41.1 & 76.3 & 19.6 & 56.2 & 4.41 & 15.8 & 12.5 & 25.0 & 19.4 & 43.3 \\
            \quad GRPO          & 44.2 & 72.1 & 25.8 & 53.6 & 5.88 & 18.8 & 10.0 & \underline{42.5} & 21.5 & 46.8 \\
            \multicolumn{11}{l}{\textbf{Training-Free Latent Reasoning}} \\
            \quad Soft Thinking (Gumbel)    & 41.3 & 77.1 & 25.4 & 54.2 & 4.78 & 19.1 & 5.00 & 32.5 & 19.1 & 45.7 \\
            \quad Soft Thinking (Dirichlet) & 38.6 & 74.2 & 24.6 & 55.4 & 2.57 & 17.3 & 12.5 & 35.0 & 19.6 & 45.5 \\
            \multicolumn{11}{l}{\textbf{RL-Trained Latent Reasoning}} \\
            \quad Soft Tokens, Hard Truths (Gaussian)    & 41.1 & 74.2 & 24.0 & 56.4 & 4.04 & 15.1 & 10.0 & 37.5 & 19.8 & 45.8 \\
            \quad Soft-GRPO (Gumbel) & 42.5 & 75.7 & 24.6 & 52.8 & 5.15 & 15.1 & 15.0 & 35.0 & 21.8 & 44.7 \\
            \quad LEPO (Gumbel)      & 42.3 & 75.1 & 25.8 & 54.4 & 4.78 & 18.4 & 15.0 & 37.5 & \cellcolor{gray!35}23.0 & 46.4 \\
            \quad DLaR               & \underline{50.9} & \underline{80.7} & \underline{30.0} & \underline{59.6} & \underline{5.88} & \underline{19.5} & \underline{25.0} & \underline{42.5} & \textbf{27.9} & \textbf{50.6} \\
            \quad \quad w/o TAST     & 43.4 & 74.8 & 23.2 & 55.6 & 5.15 & 18.8 & 12.5 & 30.0 & 21.1 & 44.8 \\
            \quad \quad w/o ASM      & 44.9 & 75.9 & 25.8 & 56.0 & 5.51 & \underline{19.5} & 10.0 & \underline{42.5} & 22.8 & \cellcolor{gray!35}48.5 \\
            \hline
            \multicolumn{11}{l}{\cellcolor{cyan!20}\textbf{Qwen2.5-1.5B}} \\
            \hline
            \multicolumn{11}{l}{\textbf{Discrete Reasoning}} \\
            \quad Base Model     & 53.8 & 89.5 & 40.0 & 73.2 & 9.19 & 29.4 & 25.0 & 40.0 & 32.0 & 58.0 \\
            \quad GRPO           & 70.9 & 91.6 & 51.4 & 76.6 & 15.4 & 34.9 & 25.0 & 50.0 & 40.7 & 63.3 \\
            \multicolumn{11}{l}{\textbf{Training-Free Latent Reasoning}} \\
            \quad Soft Thinking (Gumbel)      & 52.5 & 89.7 & 38.2 & 71.4 & 9.92 & 27.6 & 15.0 & 47.5 & 28.9 & 59.0 \\
            \quad Soft Thinking (Dirichlet)   & 53.6 & 89.9 & 37.2 & 72.6 & 10.3 & 29.0 & 20.0 & 62.5 & 30.3 & 63.5 \\
            \multicolumn{11}{l}{\textbf{RL-Trained Latent Reasoning}} \\
            \quad Soft Tokens, Hard Truths (Gaussian)   & 73.2 & 89.8 & 53.8 & 76.6 & \underline{19.5} & 33.8 & 22.5 & \underline{65.0} & 42.3 & 66.3 \\
            \quad Soft-GRPO (Gumbel)   & 74.1 & 90.4 & 54.2 & 78.4 & 16.2 & 34.9 & 30.0 & 55.0 & 43.6 & 64.7 \\
            \quad LEPO (Gumbel)        & \underline{74.2} & 90.4 & \underline{57.6} & 77.6 & 18.0 & 34.2 & 27.5 & \underline{65.0} & 44.3 & \cellcolor{gray!35}66.8 \\
            \quad DLaR              & 73.6 & \underline{92.3} & 57.4 & \underline{81.4} & 18.8 & 36.4 & \underline{37.5} & 62.5 & \textbf{46.8} & \textbf{68.2} \\
            \quad \quad w/o TAST    & 72.3 & 89.7 & 55.0 & 77.8 & 18.8 & 36.4 & 35.0 & 60.0 & \cellcolor{gray!35}45.3 & 66.0 \\
            \quad \quad w/o ASM     & 73.2 & 91.7 & 55.2 & 79.4 & 18.0 & \underline{36.8} & 22.5 & 52.5 & 42.2 & 65.1 \\
            \bottomrule
        \end{tabular}
    }
    \caption{Main results on four in-distribution mathematical reasoning benchmarks, with ablation variants of DLaR. The best result on each metric and dataset is \underline{underlined}, the best average result is \textbf{bolded}, and the second-best average result is \cellcolor{gray!20}shaded.}
    \label{tab:main_results}
\end{table*}

\textbf{Main Results.}
Table~\ref{tab:main_results} reports the performance of all methods on four in-distribution mathematical reasoning benchmarks. DLaR achieves the highest average across both backbones: 27.9\% Pass@1 / 50.6\% Pass@8 on Llama-3.2-1B, and 46.8\% / 68.2\% on Qwen2.5-1.5B, with pronounced gains on GSM8K (50.9\% vs. LEPO's 42.3\% Pass@1) and MATH500 (30.0\% vs. 25.8\%).

We make three observations. First, on the 1B backbone, all stochastic latent reasoning methods achieve higher Pass@8 than GRPO on most benchmarks, confirming that stochasticity enhances exploratory capacity. Second, Soft Thinking (Gumbel) achieves higher Pass@8 on GSM8K but drops substantially on MinervaMath, indicating that without training, fixed stochasticity cannot reliably preserve reasoning coherence. Third, on the 1B backbone, fixed-stochasticity RL-trained methods underperform GRPO on Pass@1 despite higher Pass@8—a pattern mitigated on Qwen2.5-1.5B, which has undergone mathematical pretraining. While individual baselines achieve the best performance on specific benchmark-metric combinations, DLaR achieves the highest average across all benchmarks and both backbones, demonstrating that adaptive stochasticity control remains beneficial regardless of pretraining.

\textbf{Ablation Study.}
We conduct two ablations to isolate the contribution of each mechanism (Table~\ref{tab:main_results}). On the 1B backbone, removing TAST (\textbf{w/o TAST}) drops average Pass@1 from 27.9\% to 21.1\% and Pass@8 from 50.6\% to 44.8\%, demonstrating that the curriculum decay schedule plays a vital role in enabling effective exploration throughout training. Removing ASM (\textbf{w/o ASM}) reduces them to 22.8\% and 48.5\%, respectively, indicating that confidence-entropy fusion provides critical token-level guidance for modulating perturbation intensity. On the 1.5B backbone, the same pattern holds: both ablations underperform the full model (DLaR: 46.8/68.2; w/o TAST: 45.3/66.0; w/o ASM: 42.2/65.1). These results confirm that ASM and TAST are complementary and both essential for DLaR's effectiveness across model scales.

\subsection{Evaluation on Out-of-Distribution Benchmarks}

\begin{figure}
    \centering
    \includegraphics[width=0.98\linewidth]{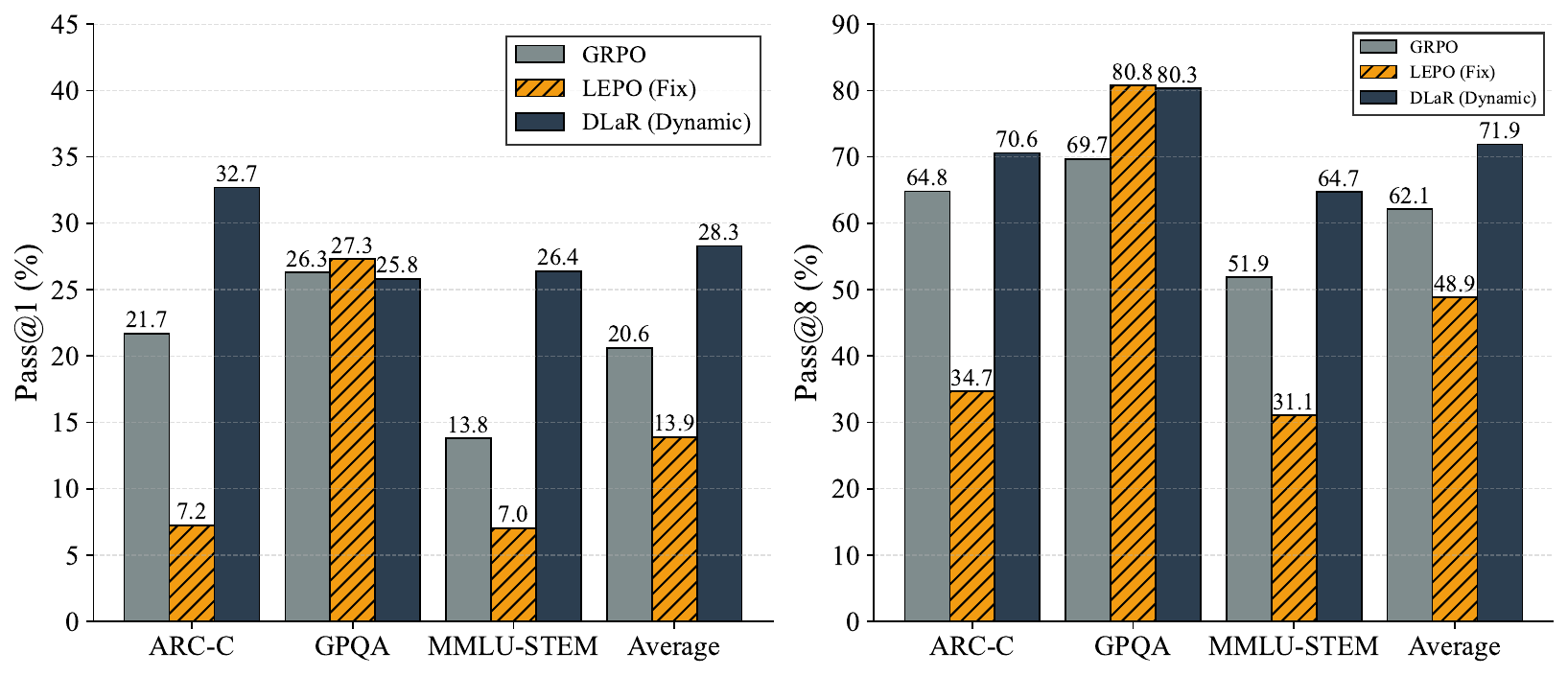}
    \caption{Out-of-distribution evaluation on ARC-C, GPQA, and MMLU-STEM. \textbf{Left:} Pass@1; \textbf{Right:} Pass@8.}
    \label{fig:ood_results}
\end{figure}

To assess generalization, we evaluate on ARC-C, GPQA, and MMLU-STEM. As shown in Figure~\ref{fig:ood_results}, DLaR substantially outperforms both baselines on ARC-C and MMLU-STEM, while performing comparably to LEPO on GPQA. Unlike GSM8K and MATH, which require multi-step reasoning with diverse solution paths, GPQA is a graduate-level multiple-choice benchmark where each question has a single correct answer and a relatively deterministic reasoning path, making adaptive exploration less beneficial. Despite this, DLaR achieves the highest average across all three benchmarks, confirming that its overall generalization is superior while its benefits are most pronounced on reasoning-intensive tasks.

\subsection{Latent Space Shift Analysis}
\label{subsec:latent_shift}

\begin{figure*}
    \centering
    \includegraphics[width=0.98\linewidth]{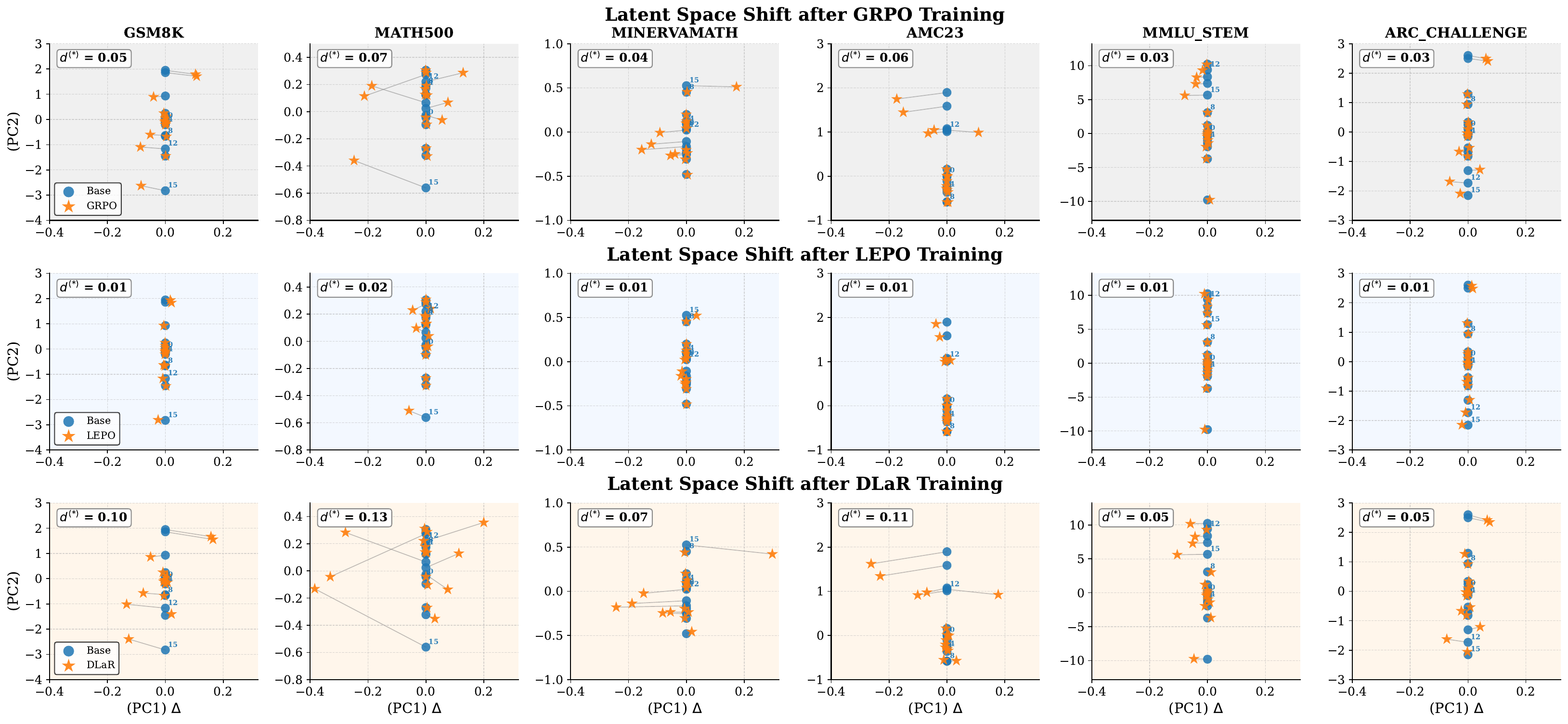}
    \caption{Latent space shift of Llama-3.2-1B across six datasets. Gray lines connect the same layer before and after training. $d^{(*)}$ denotes the average Euclidean distance between pre- and post-training centroids.}
    \label{fig:latent_shift}
\end{figure*}

To quantify how adaptive stochasticity control (DLaR) reshapes internal representations compared to fixed stochasticity (LEPO), we measure the per-layer Euclidean distance between the trained and initialized centroids \citep{zheng2025spurious,xu2026unlearning,huan2026does}, $d_{l,\mathcal{D}} = \| \mathbf{z}_{l,\mathcal{D}}^{\text{(trained)}} - \mathbf{z}_{l,\mathcal{D}}^{\text{(init)}} \|_2$, and average it across layers to obtain $d^{(*)}$. Figure~\ref{fig:latent_shift} visualizes the per-layer trajectories of GRPO, LEPO, and DLaR, and Table~\ref{tab:pca_shift} summarizes the aggregate $d^{(*)}$ across Math and Knowledge datasets.

\begin{table}[!ht]
    \centering
    \begin{tabular}{l c c}
        \toprule
        Model & Math Datasets & Knowledge Datasets \\
        \midrule
        GRPO & 0.057 $\pm$ 0.010 & 0.030 $\pm$ 0.002 \\
        LEPO & 0.012 $\pm$ 0.004 & 0.012 $\pm$ 0.002 \\
        DLaR & \textbf{0.103 $\pm$ 0.021} & \textbf{0.049 $\pm$ 0.002} \\
        \bottomrule
    \end{tabular}
    \caption{PCA shift distances of representation centers across models.}
    \label{tab:pca_shift}
\end{table}

Three observations stand out. \textbf{First}, DLaR exhibits substantially larger representational drift than both LEPO and GRPO across all datasets ($8.6\times$ and $1.8\times$ on Math, respectively). \textbf{Second}, this pattern is consistent across mathematical and knowledge-intensive datasets, indicating that the larger drift is a general property of DLaR's training dynamics rather than a dataset-specific artifact. \textbf{Third}, DLaR shares a similar depth-dependent pattern with GRPO---both exhibit monotonically increasing drift from shallow to deep layers---whereas LEPO maintains an unusually small and nearly flat drift across all layers.

We emphasize that this \textbf{representational drift} differs fundamentally from the \textbf{semantic drift} in Section~\ref{sec:intro}: the latter pushes representations \textit{off the semantic manifold} and breaks reasoning coherence (Figure~\ref{fig:noise_compare}), whereas the former is a \textit{directed on-manifold reorganization} guided by the confidence-entropy signal. Notably, DLaR's drift pattern closely mirrors GRPO's, while LEPO's flat drift reflects a near-initialization regime. We therefore interpret DLaR's larger drift as evidence that adaptive stochasticity control permits the model to reorganize its representation to a degree that fixed stochasticity cannot achieve.

\subsection{Hyperparameter Analysis}
\label{app:hyperparameter}

\begin{table}[!ht]
    \centering
    \small
    \setlength{\tabcolsep}{4pt}
    \resizebox{\linewidth}{!}{
        \begin{tabular}{l c c c c}
            \toprule[1.5pt]
            \textbf{Config} & \multicolumn{2}{c}{\textbf{GSM8K}} & \multicolumn{2}{c}{\textbf{Avg (4 datasets)}} \\
            \cmidrule(lr){2-3} \cmidrule(lr){4-5}
            ($\alpha$ / Schedule / $\tau_{\text{start}}$) & Pass@1 & Pass@8 & Pass@1 & Pass@8 \\
            \midrule[1.5pt]
            \multicolumn{5}{l}{\textit{Confidence Weight ($\alpha$)}} \\
            \quad $0.1$ / Linear / $0.25$ & 49.9 & 76.9 & 22.6 & 47.2 \\
            \quad $0.3$ / Linear / $0.25$ & 49.9 & 76.0 & 25.4 & 45.9 \\
            \quad $0.5$ / Linear / $0.25$ & 49.9 & 80.1 & 25.4 & 49.2 \\
            \quad $0.7$ / Linear / $0.25$ & \textbf{51.1} & \textbf{80.7} & \textbf{25.6} & \textbf{50.0} \\
            \quad $0.9$ / Linear / $0.25$ & 49.9 & 79.8 & 24.9 & 48.7 \\
            \midrule
            \multicolumn{5}{l}{\textit{Stochasticity Tuning Schedule}} \\
            \quad $0.7$ / Linear / $0.25$ & 51.1 & \textbf{80.7} & 25.6 & \textbf{50.0} \\
            \quad $0.7$ / Cosine / $0.25$ & 51.3 & 76.0 & 24.5 & 47.7 \\
            \quad $0.7$ / Step / $0.25$   & \textbf{51.8} & 76.5 & \textbf{25.8} & 47.0 \\
            \midrule
            \multicolumn{5}{l}{\textit{Initial Base Stochasticity ($\tau_{\text{start}}$)}} \\
            \quad $0.7$ / Linear / $0.25$ & 51.1 & 80.7 & 25.6 & 50.0 \\
            \quad $0.7$ / Linear / $0.35$ & 50.9 & \textbf{80.7} & \textbf{27.9} & \textbf{50.6} \\
            \quad $0.7$ / Linear / $0.45$ & \textbf{51.8} & 75.9 & 25.5 & 45.6 \\
            \bottomrule[1.5pt]
        \end{tabular}
    }
    \caption{Hyperparameter ablation study on GSM8K and averaged across four in-distribution benchmarks. Best results are in \textbf{bold}.}
    \label{tab:hyperparameter_ablation}
\end{table}

We analyze three key hyperparameters using the Llama-3.2-1B backbone, and validate the optimal configuration across all four in-distribution benchmarks. Table~\ref{tab:hyperparameter_ablation} summarizes the results. Based on these findings, we adopt $\alpha = 0.7$, Linear Tuning, and $\tau_{\text{start}} = 0.35$ as the default configuration for DLaR.

\textbf{Confidence Weight.}
For the confidence weight $\alpha$, we compare $\alpha \in \{0.1, 0.3, 0.5, 0.7, 0.9\}$. The configuration $\alpha = 0.7$ achieves the best performance on both GSM8K and the 4-dataset average, consistently outperforming other values. Lower values such as $\alpha = 0.1$ and $\alpha = 0.3$ perform notably worse, confirming that confidence should serve as the primary signal while entropy provides complementary information.

\textbf{Stochasticity Tuning Schedule.}
We compare the three tuning schedules introduced in Section~\ref{subsec:tast}. Linear Tuning achieves the best Pass@8 on both GSM8K (80.7\%) and the 4-dataset average (50.0\%), while Cosine and Step Tuning achieve slightly higher GSM8K Pass@1 (51.3\% and 51.8\% vs.\ 51.1\%). We adopt Linear Tuning as the default, as it provides the most consistent performance across metrics and benchmarks. 

\textbf{Initial Base Stochasticity.}
We evaluate three values of $\tau_{\text{start}}$ (0.25, 0.35, and 0.45). On GSM8K, $\tau_{\text{start}} = 0.45$ achieves the best Pass@1 (51.8\%), while $\tau_{\text{start}} = 0.25$ and $\tau_{\text{start}} = 0.35$ tie for the best Pass@8 (80.7\%). When averaged across all four benchmarks, $\tau_{\text{start}} = 0.35$ achieves the best performance on both metrics (27.9\% Pass@1, 50.6\% Pass@8), outperforming $\tau_{\text{start}} = 0.25$ (25.6\% / 50.0\%) and $\tau_{\text{start}} = 0.45$ (25.5\% / 45.6\%). We therefore adopt $\tau_{\text{start}} = 0.35$ as the default, as it provides the best cross-dataset generalization while remaining competitive on GSM8K.

\subsection{Case Study}
\label{app:case_study}

\begin{figure*}
    \centering
    \includegraphics[width=\linewidth]{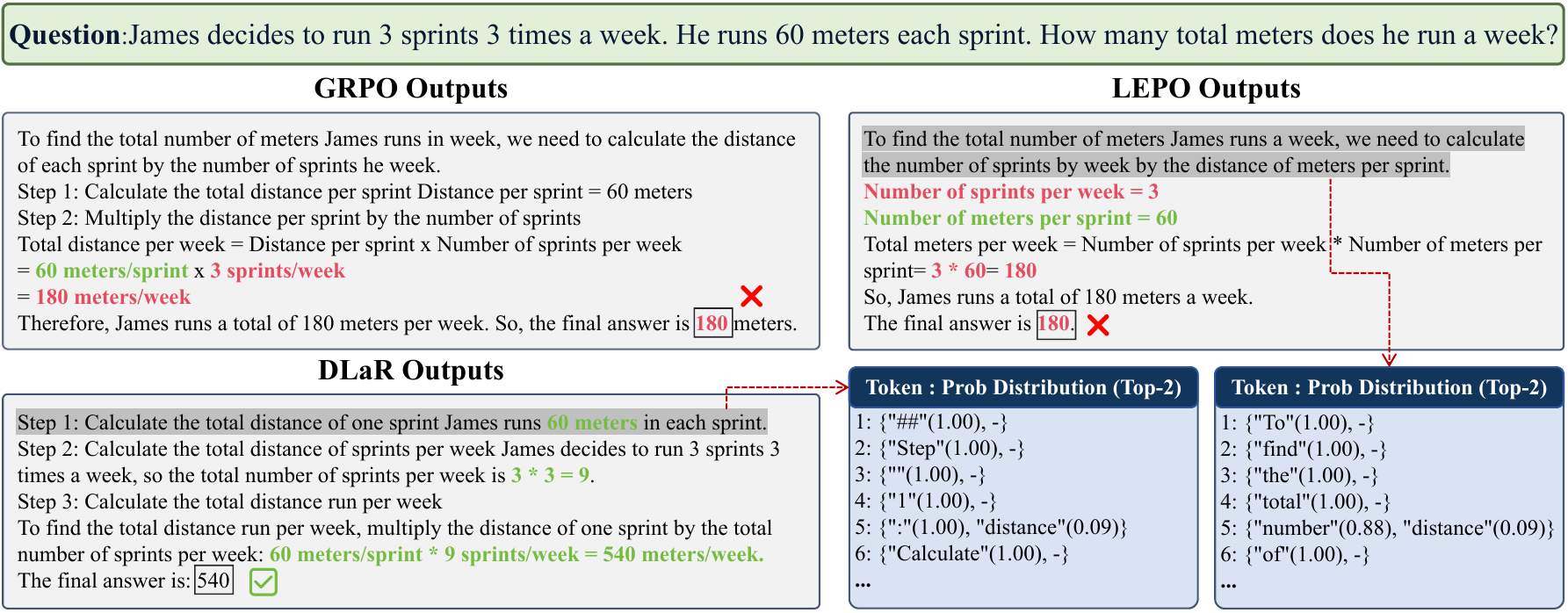}
    \caption{Example from GSM8K illustrating common failure patterns of fixed stochasticity baselines and the behavior of DLaR.}
    \label{fig:case_study}
\end{figure*}

Figure~\ref{fig:case_study} illustrates a representative example from GSM8K that highlights typical failure patterns of fixed stochasticity baselines. Both GRPO and LEPO miss the key multiplier ``3 times per week'' and compute \(60 \times 3 = 180\). However, their failure modes differ: GRPO produces disorganized reasoning with unclear intermediate steps, while LEPO follows a clearer logical structure but still omits the critical multiplier. In contrast, DLaR correctly identifies the total number of sprints as \(3 \times 3 = 9\), yielding \(60 \times 9 = 540\), and generates coherent step-by-step reasoning. This case suggests that DLaR's confidence-entropy guided modulation helps maintain reasoning coherence and mitigates the omissions often introduced by fixed stochasticity.

\subsection{Training Dynamics}
\label{app:training_dynamics}

\begin{figure*}
    \centering
    \begin{subfigure}[b]{0.48\linewidth}
        \centering
        \includegraphics[width=\textwidth]{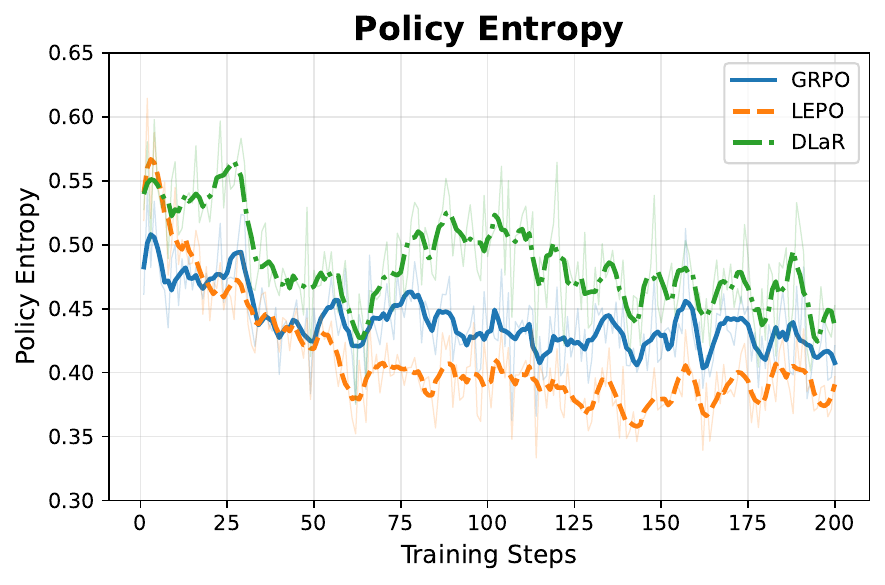}
        \caption{}
        \label{fig:training_entropy}
    \end{subfigure}
    \hfill
    \begin{subfigure}[b]{0.48\linewidth}
        \centering
        \includegraphics[width=\textwidth]{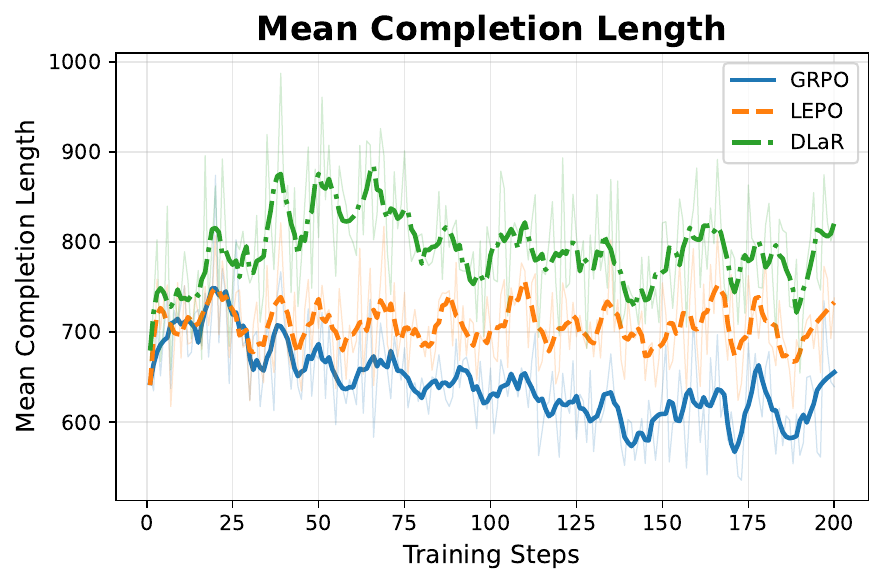}
        \caption{}
        \label{fig:training_length}
    \end{subfigure}
    \\[0.5em]
    \begin{subfigure}[b]{0.48\linewidth}
        \centering
        \includegraphics[width=\textwidth]{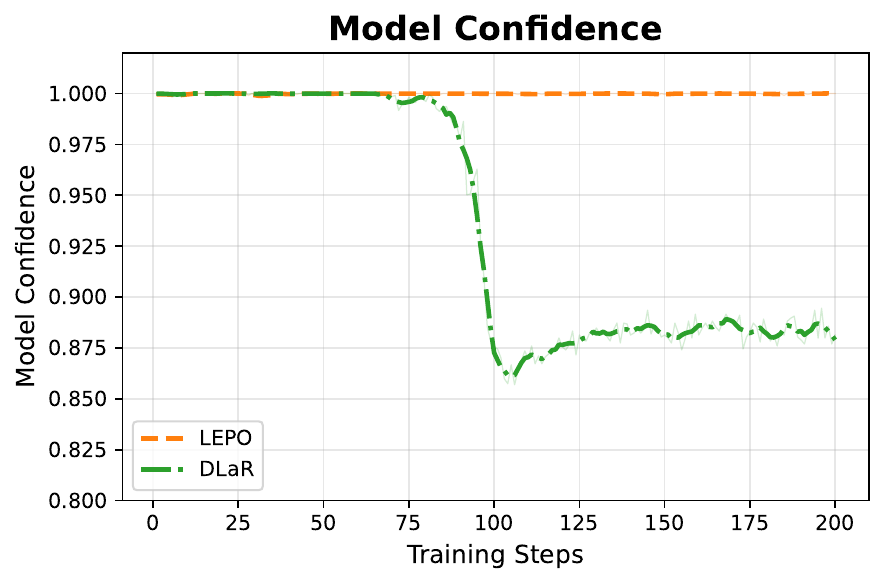}
        \caption{}
        \label{fig:training_confidence}
    \end{subfigure}
    \hfill
    \begin{subfigure}[b]{0.48\linewidth}
        \centering
        \includegraphics[width=\textwidth]{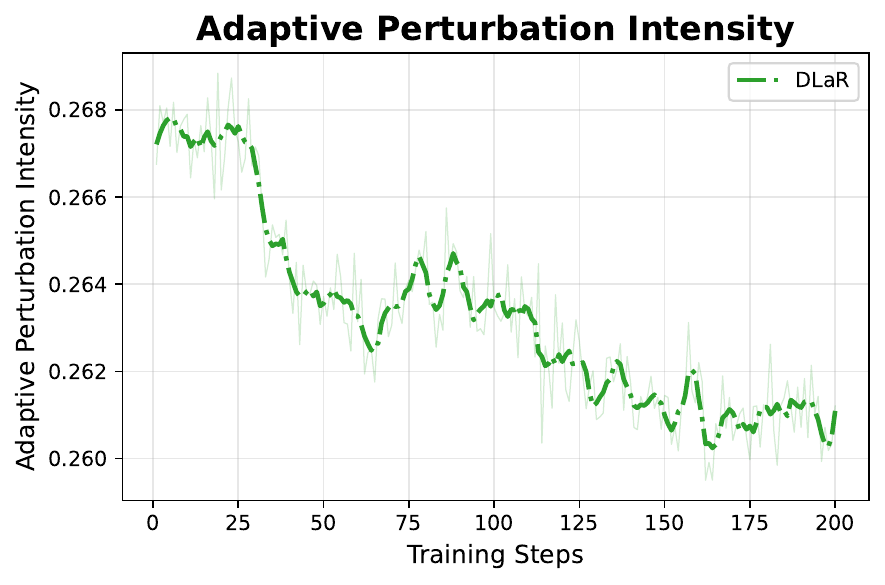}
        \caption{}
        \label{fig:training_adaptive_perturbation}
    \end{subfigure}
    \caption{Training dynamics on GSM8K with Llama-3.2-1B. DLaR maintains higher policy entropy throughout training, while its adaptive perturbation intensity decreases smoothly following the curriculum schedule.}
    \label{fig:training_dynamics}
\end{figure*}

To understand how DLaR's adaptive stochasticity control shapes the learning process, we analyze the training dynamics of DLaR, LEPO, and GRPO on GSM8K with the Llama-3.2-1B backbone. Figure~\ref{fig:training_dynamics} presents four key metrics: policy entropy, mean completion length, model confidence, and adaptive perturbation intensity.

\textbf{Policy Entropy.}
DLaR maintains consistently higher entropy than GRPO and LEPO throughout training (Figure~\ref{fig:training_entropy}). GRPO's entropy drops rapidly, indicating premature convergence to deterministic policies, while LEPO's entropy declines steadily as fixed noise cannot prevent policy collapse. DLaR's elevated and stable entropy confirms that adaptive stochasticity control enables sustained exploration: ASM increases perturbation at uncertain steps, while TAST reduces base stochasticity for later exploitation.

\textbf{Mean Completion Length.}
DLaR generates the longest solutions, GRPO the shortest, and LEPO lies in between (Figure~\ref{fig:training_length}). Longer completions reflect more diverse reasoning paths, consistent with the higher entropy observed above. GRPO's shorter completions suggest that without stochastic exploration, the model commits to the most direct path and terminates early.

\textbf{Model Confidence.}
LEPO maintains consistently high confidence throughout training, whereas DLaR stays at similarly high confidence early on and then declines to $\sim$0.875 after around step 70 (Figure~\ref{fig:training_confidence}). This divergence reflects ASM's effect: as TAST reduces the base stochasticity, ASM increasingly concentrates perturbation on uncertain steps, preventing uniform overconfidence and encouraging the model to remain calibrated to genuine uncertainty. DLaR's moderate confidence thus indicates a healthier exploration-exploitation balance than LEPO's fixed overconfidence.

\textbf{Adaptive Perturbation Intensity.}
The adaptive perturbation intensity in DLaR decreases smoothly over training (Figure~\ref{fig:training_adaptive_perturbation}), following the TAST curriculum schedule—high early for broad exploration, decaying later for fine-grained exploitation. This demonstrates that ASM and TAST work in concert: TAST provides the global trend, while ASM modulates step-wise intensity based on confidence and entropy.

\subsection{Computational Cost Analysis}

\begin{table*}[!ht]
    \centering
    \small
    \setlength{\tabcolsep}{3pt}
    \begin{tabular}{l c c c c c c c c}
        \toprule
        \multirow{3}{*}{Method} 
            & \multicolumn{4}{c}{Inference} 
            & \multicolumn{3}{c}{Training} 
            & \multirow{3}{*}{Acc. (\%)} \\
        \cmidrule(lr){2-5} \cmidrule(lr){6-8}
            & Latency & Throughput & Lat./Sample & Avg. Len 
            & Time & Mem & Rollout \\
            & (ms/tok) & (tok/s) & (s) & (tok) 
            & (h) & (GB) & Overhead & \\
        \midrule
        GRPO & 14.19 & 70.48 & 3.156 & 222.40 & $\sim$38.3 & 27.9 & -- & 44.2 \\
        LEPO & 14.43 & 69.32 & 2.786 & 193.10 & $\sim$40.4 & 27.8 & +5.5\% / $-$0.1\% & 42.3 \\
        DLaR & 14.29 & 70.00 & 3.198 & 223.86 & $\sim$41.7 & 29.8 & +8.9\% / +7.0\% & 50.9 \\
        \bottomrule
    \end{tabular}
    \caption{Computational cost of DLaR compared to GRPO and LEPO on GSM8K (Llama-3.2-1B). Rollout overhead reports wall-clock time / peak memory relative to GRPO.}
    \label{tab:cost}
\end{table*}

We analyze the computational overhead of DLaR's adaptive stochasticity control. DLaR introduces confidence and entropy estimation at each latent step, which reuses the same logits already computed for token sampling and adds no extra forward passes. As shown in Table~\ref{tab:cost}, per-token latency is nearly identical across all methods (14.19--14.43 ms/tok), confirming that ASM and TAST are confined to the rollout stage and incur no inference overhead. Consequently, differences in per-sample latency stem entirely from sequence length: LEPO produces the shortest solutions (193.10 tokens), while GRPO and DLaR generate comparable lengths (222.40 vs.\ 223.86) with similar per-sample latencies (3.156 vs.\ 3.198\,s). Notably, at comparable generation length, DLaR outperforms GRPO by 6.7 points, indicating that the gain arises from adaptive stochasticity rather than longer reasoning chains.

During training, DLaR incurs 8.9\% additional wall-clock time and 7.0\% additional peak memory over GRPO. Since LEPO shares the latent architecture but omits adaptive control, this memory increase originates from ASM and TAST rather than the latent architecture itself. Given the modest overhead and substantial accuracy gain, DLaR provides a favorable cost-performance trade-off. All inference experiments are conducted on a single NVIDIA A100 GPU with 40GB memory, using a batch size of 1. Latency is measured after 5 warmup iterations with CUDA synchronization enabled. Training cost is measured on 2$\times$A100-40GB GPUs with DeepSpeed ZeRO-2.

\section{Conclusion}
\label{sec:conclusion}

In this work, we identified two limitations of existing stochastic latent reasoning methods: uniform stochasticity across all steps is suboptimal, and unconstrained perturbations disrupt reasoning coherence. To address them, we proposed \textbf{DLaR}, a framework that introduces dynamic stochasticity control within GRPO through two complementary mechanisms: \textbf{ASM}, which adjusts perturbation intensity at each latent step based on a confidence-entropy fusion signal, and \textbf{TAST}, which progressively decays the base stochasticity level following a curriculum schedule. Extensive experiments on in-distribution and out-of-distribution mathematical reasoning benchmarks demonstrate that DLaR achieves the highest average performance across model scales. Our analysis further reveals that DLaR maintains appropriate entropy levels throughout training, enabling sustained exploration without compromising reasoning quality.

\section*{Limitations}

Our work has several limitations. First, DLaR is currently evaluated only on mathematical reasoning benchmarks and models up to 1.5B parameters; whether the observed benefits generalize to larger models (e.g., 7B and beyond) and to other reasoning domains such as code generation, scientific reasoning, or commonsense question answering remains an open question. Second, our hyperparameter analysis focuses on the initial base stochasticity $\tau_{\text{start}}$, the confidence weight $\alpha$, and the stochasticity tuning schedule; a principled method for automatically selecting these hyperparameters across tasks is left for future work. Third, our results on GPQA suggest that adaptive stochasticity control is most beneficial when the task requires exploring diverse reasoning trajectories, and its advantage is less pronounced on knowledge-intensive benchmarks where the reasoning path is relatively deterministic; understanding what task characteristics determine the magnitude of DLaR's gains requires further investigation.

\section*{Acknowledgments}

This work was supported by the National Natural Science Foundation of China (Grant Nos. 62537001 and 62276179) and the Project Funded by the Priority Academic Program Development of Jiangsu Higher Education Institutions.

\printcredits

\bibliographystyle{model5-names}
\bibliography{cas-refs}

\appendix

\section{Datasets Details}
\label{appendix:datasets}

\begin{table}[!ht]
    \centering
    \small
    \resizebox{\linewidth}{!}{
        \begin{tabular}{l l c c}
            \toprule
            \textbf{Dataset} & \textbf{Domain} & \textbf{Size} & \textbf{Task Type} \\
            \midrule
            DAPO-Math-17k & Mathematics & 17,000 & Training \\
            \midrule
            GSM8K & Grade school math & 1,319 & Numerical answer \\
            MATH500 & Competition math & 500 & Numerical answer \\
            MinervaMath & STEM & 272 & Numerical answer \\
            AMC23 & Competition math & 40 & Numerical answer \\
            \midrule
            ARC-C & Science & 1,172 & Multiple choice \\
            GPQA & Graduate-level STEM & 198 & Multiple choice \\
            MMLU-STEM & STEM & 3,000 & Multiple choice \\
            \bottomrule
        \end{tabular}
    }
    \caption{Statistics of the training and evaluation datasets.}
    \label{tab:dataset_stats}
\end{table}

We organize this section into three parts: the training dataset, the in-distribution evaluation datasets, and the out-of-distribution evaluation datasets. Table~\ref{tab:dataset_stats} summarizes the statistics of all datasets used in our experiments.

\subsection{Training Dataset}

\textbf{DAPO-Math-17k} \citep{yu2026dapo} is a large-scale mathematical reasoning dataset containing approximately 17,000 high-quality problems with verifiable numerical answers. It covers a wide range of mathematical topics and difficulty levels, and is specifically designed for reinforcement learning with verifiable rewards (RLVR). All models in our experiments are trained on this dataset.

\subsection{In-Distribution Evaluation Datasets}

\textbf{GSM8K} \citep{cobbe2021training} consists of 1,319 grade school math word problems, each requiring 2--8 reasoning steps to solve. We use the official test split containing 1,319 problems.

\textbf{MATH500} \citep{hendrycks2021measuring} is a 500-problem subset of the MATH dataset, covering five domains: algebra, combinatorics, geometry, number theory, and precalculus. Each problem requires multi-step reasoning and is significantly more challenging than GSM8K.

\textbf{MinervaMath} \citep{lewkowycz2022minerva} contains 272 problems from STEM subjects, including physics, chemistry, and mathematics, curated from undergraduate-level courses. The problems require quantitative reasoning and precise numerical answers.

\textbf{AMC23} \citep{he2026deepmath} consists of 40 problems from the 2023 American Mathematics Competitions (AMC), covering algebra, geometry, and combinatorics. These problems are designed to test mathematical insight and creative problem-solving.

\subsection{Out-of-Distribution Evaluation Datasets}

\textbf{ARC-C} \citep{clark2018think} contains 1,172 multiple-choice science questions at the grade-school level, requiring reasoning over scientific knowledge. We use the Challenge Set (ARC-C) split.

\textbf{GPQA} \citep{rein2024gpqa} is a graduate-level multiple-choice question-answering benchmark containing 448 questions across biology, chemistry, and physics. The questions are designed to be ``Google-proof''—difficult for non-experts to answer even with web access. We use the Diamond subset containing 198 questions.

\textbf{MMLU-STEM} \citep{hendrycks2021measuring} is a subset of the MMLU benchmark covering STEM subjects, including mathematics, physics, chemistry, computer science, and engineering. We evaluate on the STEM portion containing approximately 3,000 multiple-choice questions.

\section{More Experimental Details}
\label{app:experimental}

\subsection{Prompt Templates}
\label{app:prompts}

We describe the prompt templates used for evaluation. The training prompt is identical to the mathematical reasoning prompt described below.

\begin{figure}
    \centering
    \includegraphics[width=0.95\linewidth]{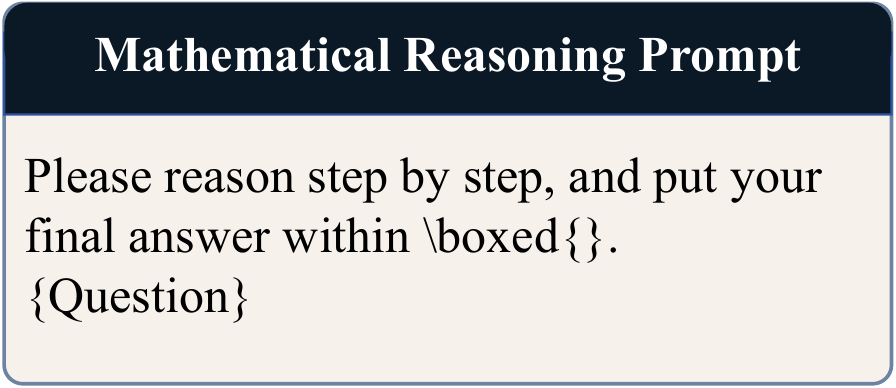}
    \caption{Mathematical reasoning prompt template. The same template is used for RL training on DAPO-Math-17k.}
    \label{fig:prompt_math}
\end{figure}

\textbf{Mathematical Reasoning Prompt.}
For in-distribution mathematical reasoning benchmarks (GSM8K, MATH500, MinervaMath, and AMC23), we use the evaluation prompt template shown in Figure~\ref{fig:prompt_math}.

\begin{figure}
    \centering
    \includegraphics[width=0.95\linewidth]{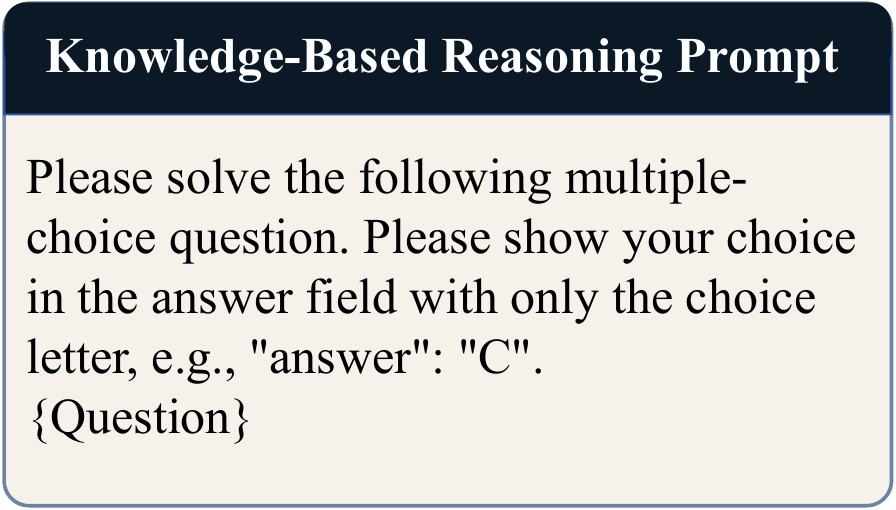}
    \caption{Knowledge-based reasoning prompt template.}
    \label{fig:prompt_knowledge}
\end{figure}

\textbf{Knowledge-Based Reasoning Prompt.}
For the knowledge-based reasoning benchmarks (ARC-C, GPQA, and MMLU-STEM), we use the multiple-choice evaluation prompt template shown in Figure~\ref{fig:prompt_knowledge}.

\subsection{More Implementation Details}
\label{app:implementation_details}

\textbf{Training.} All models are trained for one epoch (approximately 1 day for Llama-3.2-1B and 1.5 days for Qwen2.5-1.5B) with the cosine learning rate scheduler, a warmup ratio of 0.03, and a maximum gradient norm of 1.0. We use a learning rate of $1 \times 10^{-6}$ with AdamW ($\beta_1 = 0.9$, $\beta_2 = 0.99$) and a weight decay of 0.1. The per-device batch size is 1 with a gradient accumulation of 256 steps. The base stochasticity level $\tau_{\text{base}}^{(s)}$ follows Linear Tuning (as defined in Section~\ref{subsec:tast}), decaying from $\tau_{\text{start}} = 0.35$ to $\tau_{\text{end}} = 0.02$ over $S_{\text{decay}} = 1000$ steps. Training uses a maximum sequence length of 2048 tokens and a maximum prompt length of 1024 tokens. All training runs use a fixed random seed of 42.

\textbf{Latent and Noise Configuration.} We set the latent length to $L_g = 32$, the adaptation strength to $\lambda = 0.5$, and the confidence weight to $\alpha = 0.7$ (with entropy weight $1 - \alpha = 0.3$). The final perturbation intensity is clipped to $[0.01, 0.3]$ to prevent both deterministic collapse and excessive semantic drift.

\textbf{Inference.} All methods are evaluated under two decoding configurations: (1) standard discrete decoding with temperature 0.6, top-$p = 0.95$, and top-$k = 30$; and (2) Soft Thinking with Gumbel-Softmax at $\tau_g = 0.5$. We report the best result for each method. All evaluation runs use a fixed random seed of 0.

\textbf{Infrastructure.} All experiments are conducted on two NVIDIA A100 GPUs with 40GB memory each, using DeepSpeed ZeRO-2.

\subsection{Baseline Configurations}
\label{appendix:baselines}

All baseline methods are reproduced using their official codebases with default hyperparameters. 
For \textbf{LEPO}, we adopt Gumbel-Softmax with temperature $\tau_g=0.5$, where the soft token is sampled as $\mathbf{z}_{t,k} = \exp((\log \pi_{t,k} + \epsilon_k)/\tau_g) / \sum_j \exp((\log \pi_{t,j} + \epsilon_j)/\tau_g)$ with $\epsilon_k \sim \text{Gumbel}(0,1)$. 
For \textbf{Soft Tokens, Hard Truths}, we follow its official implementation and inject Gaussian noise with standard deviation $\sigma=0.05$ directly into the soft token embedding, i.e., $\hat{\mathbf{E}}_t = \mathbf{E}_t + \mathcal{N}(0, \sigma^2 I)$, where $\mathbf{E}_t$ denotes the soft token embedding defined in Equation~(\ref{eq:embedding}). 
For \textbf{Soft-GRPO}, we use its official codebase with Gumbel noise and temperature $\tau_g=0.5$. 
For the training-free \textbf{Soft Thinking} variants, we adopt Gumbel-Softmax with $\tau_g=0.5$ and Dirichlet resampling with scaling parameter $\gamma=4.0$. The latter samples token weights as $(x_1, \ldots, x_{|\mathcal{V}|}) \sim \text{Dirichlet}(\gamma \pi_t)$ and computes the soft token embedding as $\mathbf{E}_t = \sum_i x_i \cdot \mathbf{e}_i$, where $\pi_t$ is the vocabulary distribution defined in Equation~(\ref{eq:distribution}) and $\mathbf{e}_i$ is the embedding of the $i$-th vocabulary token.

\section{Algorithm Overview}

The complete training procedure is summarized in Algorithm~\ref{alg:dlr}.

\begin{algorithm*}[!ht]
    \caption{DLaR: Dynamic Stochasticity Control for Latent Reasoning}
    \label{alg:dlr}
    \begin{algorithmic}[1]
        \Require Policy model $\pi_\theta$, reference model $\pi_{\text{ref}}$, dataset $\mathcal{D}$, group size $G$, decay steps $S_{\text{decay}}$, start stochasticity $\tau_{\text{start}}$, end stochasticity $\tau_{\text{end}}$, adaptation strength $\lambda$, confidence weight $\alpha$
        \State Initialize $s \leftarrow 0$, $\tau_{\text{base}}^{(0)} \leftarrow \tau_{\text{start}}$
        \For{each training step $s$}
            \State Sample batch of queries $\{q_i\}_{i=1}^B$
            \State \textbf{Rollout:}
            \For{each query $q_i$}
                \For{$j = 1$ to $G$}
                    \State Generate latent tokens with confidence-entropy guided perturbation:
                    \For{$t = 1$ to $L_g$}
                        \State Compute vocabulary distribution $\pi_{j,t}$
                        \State Compute confidence $c_{j,t} \leftarrow \max \pi_{j,t}$ and entropy $h_{j,t} \leftarrow -\sum \pi_{j,t} \log \pi_{j,t}$
                        \State Compute ASM signal: $\tau_{\text{ASM}}^{(t)} \leftarrow \alpha(1-c_{j,t}) + (1-\alpha)h_{j,t}/\log|\mathcal{V}|$
                        \State Compute final perturbation intensity: $\tau_{\text{noise}}^{(t,s)} \leftarrow \tau_{\text{base}}^{(s)} \cdot (1 + \lambda \cdot \tau_{\text{ASM}}^{(t)})$
                        \State Sample latent token $\mathbf{z}_{j,t} \sim \text{GumbelSoftmax}(\pi_{j,t}, \tau_{\text{noise}}^{(t,s)})$
                        \State Compute embedding $\mathbf{E}_{j,t} \leftarrow \sum_k \mathbf{z}_{j,t,k} \cdot \mathbf{e}_k$
                    \EndFor
                    \State Generate discrete answer tokens
                    \State Compute reward $r_j$
                \EndFor
            \EndFor
            \State Compute group-relative advantages $\hat{A}_j \leftarrow (r_j - \text{mean}(\mathbf{r})) / \text{std}(\mathbf{r})$
            \State \textbf{Optimization:}
            \State Compute $\mathcal{J}_{\text{DLaR}}(\theta)$ via Equation~(\ref{eq:dlr_objective})
            \State Update $\theta$ via gradient descent
            \State \textbf{Update base stochasticity level:}
            \State $\tau_{\text{base}}^{(s+1)} \leftarrow \text{DecaySchedule}(\tau_{\text{start}}, \tau_{\text{end}}, s, S_{\text{decay}})$
        \EndFor
    \end{algorithmic}
\end{algorithm*}

\end{document}